%% file: fedcsis_2017.tex
\documentclass[a4paper,conference]{IEEEtran}

\usepackage[OT4,T1]{fontenc}

\usepackage{balance}

\IEEEoverridecommandlockouts 
\usepackage{graphicx}

\usepackage{algorithmic}
\usepackage{algorithm}
\usepackage{listings}
\usepackage[OT4,T1]{fontenc}
\usepackage[cmex10]{amsmath}
\usepackage{url}
\usepackage{multirow}
\usepackage{mathtools}
\usepackage{amsfonts,amssymb}
\usepackage[caption=false,font=footnotesize]{subfig}
\usepackage{stackengine}
\stackMath
\newlength\matfield
\newlength\tmplength
\def\matscale{1.}
\newcommand\dimbox[3]{%
  \setlength\matfield{\matscale\baselineskip}%
  \setbox0=\hbox{\vphantom{X}\smash{#3}}%
  \setlength{\tmplength}{#1\matfield-\ht0-\dp0}%
  \fboxrule=1pt\fboxsep=-\fboxrule\relax%
  \fbox{\makebox[#2\matfield]{\addstackgap[.5\tmplength]{\box0}}}%
}
\newcommand\raiserows[2]{%
   \setlength\matfield{\matscale\baselineskip}%
   \raisebox{#1\matfield}{#2}%
}
\newcommand\matbox[5]{
  \stackunder{\dimbox{#1}{#2}{$#5$}}{\scriptstyle(#3\times #4)}%
}
\title{Improving text classification with vectors of reduced precision\thanks{This research was supported in part by Faculty of Management and Social Communication of the Jagiellonian University and PLGrid Infrastructure.}}
\author{\IEEEauthorblockN{Krzysztof Wr\'{o}bel\IEEEauthorrefmark{1}\IEEEauthorrefmark{2},
Maciej Wielgosz\IEEEauthorrefmark{2}\IEEEauthorrefmark{3},
Marcin Pietro\'{n}\IEEEauthorrefmark{2}\IEEEauthorrefmark{3},
Micha\l{} Karwatowski\IEEEauthorrefmark{2}\IEEEauthorrefmark{3}, 
Aleksander Smywi\'{n}ski-Pohl\IEEEauthorrefmark{2}}
\IEEEauthorblockA{\IEEEauthorrefmark{1}Jagiellonian University\\
Krak\'{o}w, Poland}
\IEEEauthorblockA{\IEEEauthorrefmark{2}AGH University of Science and Technology\\
Krak\'{o}w, Poland\\ Email: \{kwrobel, wielgosz, pietron, mkarwat, apohllo\}@agh.edu.pl}
\IEEEauthorblockA{\IEEEauthorrefmark{3}Academic Computer Centre CYFRONET\\
Krak\'{o}w, Poland}}

\begin{document}
\maketitle              

\begin{abstract}
\input{abstract}
\end{abstract}

\input{contents}


\bibliographystyle{IEEEtran}
\bibliography{bibliography-machine-learning,bibliography-cern,bibliography-nlp,bibliography-misc,wielgosz-published,wielgosz-unpublished,bibliography}


\end{document}

%% file: abstract.tex
This paper presents the analysis of the impact of
a floating-point number precision reduction on the quality of text classification.
The precision reduction of the vectors representing the data
(e.g. TF--IDF representation in our case) allows for a decrease of computing time and
memory footprint on dedicated hardware platforms.
The impact of precision reduction on the classification quality was performed on 5 corpora,
using 4 different classifiers. Also, dimensionality reduction was taken into account.
Results indicate that the precision reduction improves classification accuracy for most cases
(up to 25\% of error reduction). In general, the reduction from 64 to 4 bits gives the best scores
and ensures that the results will not be worse than with the full floating-point representation.

%% file: contents.tex
\input{contents-00-introduction}
\input{contents-10-vector-space-model}
\input{contents-15-precision}
\input{contents-17-svd}

\input{contents-19-clf}
\input{contents-20-experiments-discussion}
\input{contents-30-conclusions-future}

%% file: contents-00-introduction.tex
\section{Introduction}
\label{section:intro}

\IEEEoverridecommandlockouts\IEEEPARstart{N}{atural} Language Processing (NLP), as well as Image Processing, is a part of
Artificial Intelligence. Despite intensive research and huge recent progress in
Deep Learning Techniques, applications of NLP have not reached a level that
would allow a construction and a practical implementation of robots and
machines
operating like humans. Such human-level solutions would allow for seamless and
smooth communication between machines and people. The future communication
interfaces will allow to convey information directly to the machines processing
units using natural language
\cite{bengio2013representation}\cite{schmidhuber2015deep}\cite{kumar2015ask}.
This future vision, however, requires a substantial progress in both speech
recognition and text processing domains. Applications of those two domains are
in an essence very similar and share most of the processing flow. In our
research \cite{karwatowskiFPGAcosinePPAM} we
focus on text processing, but the proposed modules may also be employed in
voice processing solutions.

NLP as a research and application field has been developed in a course of last
few decades
\cite{Manning1999Foundations}\cite{Collobert2011Natural}\cite{
Fitzgerald2015Semantic}\cite{Hermann2014Semantic}\cite{petrov2012Universal}.
Three different models of the language representation have been established,
namely Boolean Model, Vector Space Model (VSM) and Sparse Representation Model
\cite{Mikolov2013Efficient}. The latter model slowly becomes a standard for
applications and systems using Natural Language Processing
\cite{Mikolov2013Lingustic}. This is due to its superior performance, which in
turn results from the fact that it mimics the language representation within a
human brain \cite{hawkins2004intelligence}. It is worth noting that a language
as such belongs to a human cognition domain. It was developed by humans to
enable communication and was implemented with biological components in a neural
fashion \cite{mountcastle1997columnar}. Therefore, pure ontological models of
the language tend to be inferior to the biologically--inspired ones
\cite{hawkins2006hierarchical}.

Representation of knowledge within a human brain is highly distributed, sparse
and hierarchical \cite{hawkins2004intelligence}\cite{mountcastle1997columnar}.
Neural operations of cognition, which also involve language processing, are
performed using single bit precision. Every bit of the information carries
semantic meaning which reflects relationships between concepts acquired and
stored within the brain. Inspired by this we decided to examine to what extent
it is possible to implement such a bit processing scheme on a top of currently
used models in NLP. We focused on the Vector Space Model as one which is
popular and widely used in various applications. However, the research results
may also be transferred to the other models since all of them employ vector as a
basic representation structure. The vectors are a collection of fixed or
floating--point numbers which represent a certain dynamic range of a data
representation. It turns out that the dynamic range, at least in the case of
floating--point numbers, is too large and can accommodate much more information
than necessary. Therefore, we decided to reduce the range to the extent that, on the one
hand still preserves a required precision and on the other hand substantially decreases the
number of bits. Furthermore, such an approach also has a
beneficial effect on semantics of the processed lingual content since it leads
to a generalization. This, in turn, has all the positive properties of distributed
representations \cite{Mikolov2013Efficient}. Consequently, precision reduction
of vector representation may be perceived as way of concept generalization or
abstraction which is in its essence similar to latent semantic indexing \cite{Deerwester90indexingby}
and random projection \cite{Bingham2001Random}.


Precision reduction approach may not have significant performance impact on standard processors, as they typically operate on fixed data width, usually stored in IEEE--754 floating--point representation.
Therefore, reduction to below standard width or, moreover, not byte aligned width, does not introduce notable speedup.
The situation improves for single instruction, multiple data (SIMD) processors, like general--purpose computing on graphics processing units (GPGPU), or vector CPUs however data alignment is still required and speedup is only achieved through parallelism and reduction of clock cycles required to process given an amount of data.
Real benefits of precision reduction can be observed on fully customizable platforms, such as field-programmable gate arrays (FPGA) \cite{wielgosz2013fpga}\cite{wielgosz2013fpga2}\cite{wielgosz2012fpga2}.
They are not bound to any specific bitwidth or representation.
Data may be stored in any integer bitwidth, which can also differ between consecutive processing stages.
Narrower representation requires a less complicated circuit to execute calculations, which improves operating frequency.
Switching to fixed point representation further reduces circuit complexity, thus increases operating frequency, which can also vary between processing stages.
Data flow architecture can also be designed to process data in a parallel manner.
A combination of aforementioned features makes FPGA a very interesting choice as a hardware platform.
However, creating efficient design architecture and its implementation are not trivial and generate interesting research task.
As authors of this paper already began work on the dedicated hardware platform and presented their initial results in
\cite{intellisys2017}, we will not cover this topic. Still much effort needs to be put into FPGA implementation in order
to utilize its potential in NLP tasks.

Consequently, the paper addressed three main objectives:
\begin{itemize}
 \item an examination of the precision reduction impact on the
text classification results,
 \item proposition and practical verification of various popular classification
methods with different grade of reduced precision,
\item building a framework for precision reduction which is available on--line. 
\end{itemize}

The rest of the paper is organized as follows. Section \ref{section:vsm}
introduces vector space model as text documents representation. Section
\ref{section:precision_reduction} describe a procedure of precision reduction
used in our experiments.
Sections \ref{section:svd} and \ref{section:clf} describes SVD and parameters of the employed classifiers, respectively.
Experiments are presented in Section \ref{section:experiments}.
Finally, we present our conclusions in Section \ref{section:conclusions}.

%% file: contents-10-vector-space-model.tex
\section{Vector Space Model}
\label{section:vsm}

We decided to use Vector Space Model in our experiments to validate an adopted
approach to semantic vector reduction. The section that follows may be considered
as a short introduction to the model and its aspects that we address in the paper.

The VSM has already been successfully used as a~conventional method for the text
representation \cite{salton1975vector}.
The documents are represented as vectors in an $N$--dimensional vector space that
is built upon the $N$ different terms that occur in the considered document set
(\textit{i.e.} text corpus). Since the model ignores the order of the words it
is usually called a bag-of-words model, to reflect the proces of throwing the words
form the text to an unstructured bag.
The coefficients of the vector are the weights that identify the significance of
a particular term in the document.
Consequently, a set of documents in this scheme may be presented as the
term--document matrix.
An example of such a matrix is given in Table \ref{tab:vsm}.
\begin{table}
\small
\centering
\caption{A sample term--document matrix in the VSM}
\label{tab:vsm}
\begin{tabular}{|l|c|c|c|c|c|}
\hline
 & doc0 & doc1 & doc2 \\
\hline
term0 & 2 & 3 & 1 \\
\hline
term1 & 1 & 3 & 2 \\
\hline
\end{tabular}
\end{table}

\begin{figure}
\centering
\includegraphics[scale=0.48]{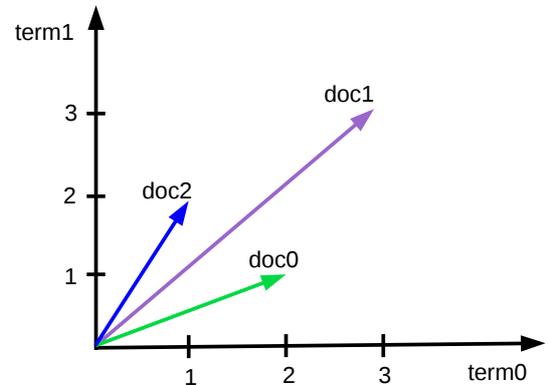}
\caption{A sample documents visualization in VSM}
\label{fig:vsm}
\end{figure}

Fig. \ref{fig:vsm} presents a simple example of the three different documents
mapped to the two--dimensional vector space, which means that they are built from
two different words.

The text similarity can be easily calculated by the cosine similarity measure in
the VSM. The cosine measure is given by the equation
\begin{equation}
 cosine \ similarity(\textbf{u},\textbf{v})=\frac{\sum_{s=0}^{N-1}(u_{s} \cdot
v_{s})}{\|\textbf{u}\| \cdot \|\textbf{v}\|},
 \label{eq:cosine}
\end{equation}
where $\textbf{v}=(v_{0},\ldots,v_{N-1})$ and
$\textbf{u}=(u_{0},\ldots,u_{N-1})$ are the vectors representing documents.

The most common method of words weighing in VSM is the computation of the
coefficients of so--called Term Frequencies (TF) and Inverted Document
Frequencies (IDF).
TF is the number of times a word appears in a given document that is normalized
\textit{i.e.} it is divided by the total number of words in a document under
consideration.
Respectively, IDF measures how common a word is among all documents in
consideration. The more common the word is, the lower its IDF is. The IDF is
computed as the ratio of
the total number of documents to the number of documents containing a given
word.

Consequently, TF--IDF is a numerical statistic that indicates how important the
word is in characterizing a given document in the context of the whole
collection.
It is often used as a weighting factor in information retrieval\footnote{It should be noted,
however that OKAPI BM25 model \cite{robertson1995okapi} that is similar to TF--IDF works better in the context
of information retrieval and is assumed a standard implemented in the major
full text search engines such as SOLR and ElasticSearch.} and text mining.
The mathematical formula for TF--IDF computation is
\begin{equation}
 tf\textnormal{-}idf_{t}=tf_{t} \cdot idf_{t}.
\end{equation}
The \textit{$tf_{t}$} value is the term \textit{$t_{t}$} frequency in the
document $d$, and it is computed as
\begin{equation}
tf_{t}=\frac{n_{t}}{\sum_{s=0}^{N-1}n_{s}},
\end{equation}
where \textit{$n_{t}$} is the number of occurrences of term \textit{$t_{t}$} in
a document $d{\in}D$.
The \textit{$idf_{t}$} value is the inverse document frequency that is given as
\begin{equation}
idf_{t}=log \frac{|D|}{|\{d\in D:t_{t} \in d \}|},
\end{equation}
where $|D|$ is the number of documents in the corpus, and $|\{d\in D:t_{t} \in d
\}|$ is the number of documents containing at least one occurrence of the term
\textit{$t_{t}$}.
The TF--IDF value increases proportionally to the number of times a word appears
in the document, but it is scaled down by the frequency of the word in the
corpus,
which helps control the fact that some words are generally more common than
others. Therefore, common words
which appear in many documents will be almost ignored. Words that appear
frequently in a single document will be scaled up.
In this work, the algorithms use the TF--IDF coefficients. 

%% file: contents-15-precision.tex
\section{Precision Reduction}
\label{section:precision_reduction}
Language models are usually very large multidimensional structures
composed of vectors. The vectors contain IEEE--754 floating--point numbers which
can be either stored in dense or sparse format for a sake of a storage space
utilization reduction. In the sparse representation only non--zero vector
elements are kept and the rest is skipped. Regardless of the representation, the
low--level elements of the vectors are numbers. In order to reduce the
representation precision, we apply a three--stage procedure given by Eq.
\ref{eq:ieee_754_single}, \ref{eq:norm_single},
\ref{eq:mapping_single}, \ref{eq:norm_single_reduced} and
\ref{eq:reduction_single} for the IEEE--754 single precision case. The procedure
for IEEE--754 double precision is the same and the equations were provided for a
sake of completeness and consistency.

We reduce precision of each vector element given by Eq. \ref{eq:ieee_754_single}
and \ref{eq:ieee_754_double}:

\begin{equation}
S_{single}: \{ \pm 2^{-126} \ldots (2 - 2^{-23}) \times 2^{127} \}^{1 \times n}
\label{eq:ieee_754_single}
\end{equation}

\begin{equation}
S_{double}: \{ \pm 2^{-1022} \ldots (2 - 2^{-52}) \times 2^{1023} \}^{1 \times
n}
\label{eq:ieee_754_double}
\end{equation}

where $S$ and $n$ is a vector of IEEE--754 floating--point numbers and its
dimension, respectively.

\begin{equation}
S_{norm-single}: \{0, \frac{1}{2^{32}} \ldots , 1 - \frac{1}{2^{32}}, 1\}^{1
\times n}
\label{eq:norm_single}
\end{equation}

\begin{equation}
S_{norm-double}: \{0, \frac{1}{2^{64}} \ldots , 1 - \frac{1}{2^{64}}, 1\}^{1
\times n}
\label{eq:norm_double}
\end{equation}

In the first step a vector of floating--point numbers: Eq.
\ref{eq:ieee_754_single} and \ref{eq:ieee_754_double} are mapped to the
fixed--point representation: Eq. \ref{eq:norm_single} and \ref{eq:norm_double} by
means of projection operation: Eq. \ref{eq:mapping_single} and
\ref{eq:mapping_double}:

\begin{equation}
S_{single} \xRightarrow{\Pi_{mapping}} S_{norm-single}
\label{eq:mapping_single}
\end{equation}

\begin{equation}
S_{double} \xRightarrow{\Pi_{mapping}} S_{norm-double}
\label{eq:mapping_double}
\end{equation}

In the second step the vector elements represented as fixed--point numbers are
mapped to the reduced representation: Eq. \ref{eq:norm_single_reduced} and
\ref{eq:norm_double_reduced} which may be regarded as a projection expressed by
Eq. \ref{eq:reduction_single} and \ref{eq:reduction_double}.

\begin{equation}
S_{single-reduced}: \{0, \frac{1}{2^{32 - r}} \ldots , 1 - \frac{1}{2^{32 - r}},
1\}^{1 \times n}
\label{eq:norm_single_reduced}
\end{equation}

\begin{equation}
S_{double-reduced}: \{0, \frac{1}{2^{64 - r}} \ldots , 1 - \frac{1}{2^{64 - r}},
1\}^{1 \times n}
\label{eq:norm_double_reduced}
\end{equation}

\begin{equation}
S_{norm-single} \xRightarrow{\Pi_{reduction}} S_{single-reduced}
\label{eq:reduction_single}
\end{equation}

\begin{equation}
S_{norm-double} \xRightarrow{\Pi_{reduction}} S_{double-reduced}
\label{eq:reduction_double}
\end{equation}

Generated TF--IDF coefficients are in IEEE--754
double floating--point format and their values span between $0$ and $1$.
Therefore to map these values to desired fixed precision is enough to multiply
them by the maximal value possible to encode with that precision.

\begin{enumerate}
  \item $max\_value = 2 ^ {bitwidth} - 1$
  \item $for\enspace tf\_idf\enspace in\enspace database:$
  \item $\qquad norm\_tf\_idf = ceil(tf\_idf * max\_value)$
\end{enumerate}

Back normalization to floating--point format is performed accordingly, only the
value needs to be divided by maximal value.

\begin{enumerate}
  \item $max\_value = 2 ^ {bitwidth} - 1$
  \item $for\enspace norm\_value\enspace in\enspace results:$
  \item $\qquad value = norm\_value / max\_value$
\end{enumerate}

The reduction parameter $r$ strongly affects performance results since it
directly decides about a number of bits which are left for the vector elements
representation. It is worth noting that it is possible to employ global
dimensionality reduction techniques such as SVD along with the methods proposed
in this paper. In this work, we consider the order of these
operations (precision reduction before or after SVD) for the sake of the best final results.

%% file: contents-17-svd.tex
\section{Singular Value Decomposition}
\label{section:svd}

Singular value decomposition (SVD) is used frequently with the term--document matrix representation.
SVD allows for dimensionality reduction of the sparse data to a low-rank dense matrix. It also addresses the
problem of polysemy in the context of text documents, since words that share meaning are mapped to vectors
occupying narrow region in the reduced vector space.

SVD is explained with the following formula:
\begin{equation}
\renewcommand\matscale{.6}
\matbox{8}{6}{m}{n}{\mathbf{M}} =
\matbox{8}{6}{m}{r}{\mathbf{U}}
\raiserows{1}{\matbox{6}{6}{r}{r}{\boldsymbol{\Sigma}}}
\raiserows{1}{\matbox{6}{6}{r}{n}{\mathbf{V}^\top}} \;,
\end{equation}
where $\mathbf{M}$ is the sparse term-document matrix, $m$ is the number of terms,
$n$ is the number of documents, $r$ is $\min(m,n)$.
In such case $\mathbf{U}$ and $\mathbf{V}$ are orthogonal matrices and
$\boldsymbol{\Sigma}$ is a diagonal matrix.

Lowering rank to $k$ is done by taking $k$ of the largest values (components)
along the diagonal of $\boldsymbol{\Sigma}$ and truncating $\mathbf{U}$ and $\mathbf{V}$.

\begin{equation}
\renewcommand\matscale{.6}
\matbox{8}{6}{m}{n}{\mathbf{M_k}} =
\matbox{8}{3}{m}{k}{\mathbf{U_k}}
\raiserows{2.5}{\matbox{3}{3}{k}{k}{\boldsymbol{\Sigma}_\mathbf{k}}}
\raiserows{2.5}{\matbox{3}{6}{k}{n}{\mathbf{V_k}^\top}}
\end{equation}

This approach is named as Latent Semantic Analysis (LSA) or Latent Semantic Indexing (LSI) \cite{deerwester1990indexing}.
A simple interpretation of LSI is that SVD combines terms into more
general concepts (synonyms or topics) and documents are represented by a weighted set of the topics.

%% file: contents-19-clf.tex
\section{Classification}
\label{section:clf}
In order to evaluate the influence of the precision reduction on the robustness of
VSM model we employed them in the problem of multi-class (single-lable) text classification.
We have chosen k--nearest neighbors algorithm (KNN), linear regression (LR) and support vector machines (SVM)
as the tested classifiers.

KNN was used with cosine similarity metric and the number of neighbors $k \in \{1,5\}$.
The algorithm does not require training, but the testing phase involves calculating similarity with every document.
It also needs to store all the documents from the training corpus. As such it is not well suited
for large corpora, which are much more popular in the recent years.

LR and SVM are similar algorithms.  In LR we applied L2 regularization.
SVM was trained with hinge loss and linear kernel.
Both execute iterative training and do not store documents for testing.

For macro--averaged objective the weights associated with classes
were adjusted inversely proportional to class frequencies in the input data
\begin{equation}
w_c = \frac{\sum n_i}{n_c},
\end{equation}
where $w_c$ is a weight associated with class $c$ and $n_i$ is a number of samples in class $i$.

%% file: contents-20-experiments-discussion.tex
\section{Experiments and the Discussion}
\label{section:experiments}

\subsection{Experimental Setup}
\label{subsection:experiments_setup}

4 modules were developed in order to execute experiments:

\subsubsection{TF--IDF}
Term frequency--inverse document frequency was calculated on training data without setting any limit on the number of words.

\subsubsection{Precision Reduction (b)}
Precision reduction was performed on VSM representation of documents as described in \ref{section:precision_reduction}, where $b$ is the precision in bits.

\subsubsection{SVD (k)}
Singular value decomposition was used to reduce the dimensionality of data, where $k$ is the number of components.

\subsubsection{Classification}
4 classifiers were used: k--nearest neighbors algorithm with cosine similarity metric for $k \in \{1,5\}$, logistic regression and support vector machines with linear kernel.

5 variants of experiments were performed:
\begin{itemize}
 \item TF--IDF and Classification,
 \item TF--IDF, Precision reduction (b) and Classification,
 \item TF--IDF, Precision reduction (b), SVD (k) and Classification,
 \item TF--IDF, SVD (k) and Classification,
 \item TF--IDF, SVD (k), Precision reduction (b) and Classification,
\end{itemize}
where $b \in \{16,8,7,6,5,4,3,2,1\}$ and $k \in \{100,200,300,400,500,1000\}$.

All results were obtained by taking an average of 5--fold cross--validation scores. Each datasets was randomly shuffled, partitioned into 5 subsets. The process of evaluation was repeated 5 times, with one subset used exactly once as testing data and the rest 4 as training data.

All experiments were performed in Python using scikit--learn \cite{scikit-learn} library with default parameters. Calculations were performed on 64--bit floating point type with 4 cores of Intel Xeon E5--2680v3.

Framework performing precision reduction is available at: \url{https://github.com/kwrobel-nlp/precision-reduction}. It determines what is the best number of bits for classification of specified corpus.
Datasets used in this work are shared for reproducibility of results.

\subsection{Datasets}
\label{subsection:dataset}

Experiments were performed on multi--class (single--label) datasets. 5 datasets are publicly available:
\begin{itemize}
 \item \texttt{webkb} - webpages collected from computer science departments,
 \item \texttt{r8} - Reuters articles with single label from R10 subcollection of Reuters-21578,
 \item \texttt{r52} - Reuters articles with single label from R90 subcollection of Reuters-21578,
 \item \texttt{20ng} - newsgroup messages,
 \item \texttt{cade} - webpages extracted from the CAD\^{E} Web Directory.
\end{itemize}

All of them are pre--processed by \cite{2007:phd-Ana-Cardoso-Cachopo}:
\begin{itemize}
 \item all letters turned to lowercase,
 \item one and two letters long words removed,
 \item stopwords removed,
 \item all words stemmed.
\end{itemize}
Multi--label datasets were transformed to single--label by removing samples with more than one class.

Table \ref{tab:corpora_stats} shows summary of corpora's main features. Corpora \texttt{webkb}, \texttt{r8}, \texttt{r52} and \texttt{20ng} are in English, \texttt{cade} is in Brazilian-Portuguese.
\texttt{cade} is the largest dataset in terms of the number of documents, vocabulary and average length of documents.
\texttt{20ng} is the most balanced (0.1 relative standard deviation), others are very skewed.

\begin{table}
\centering
\caption{Volume of datasets: number of classes, number of documents, number of unique words, average length of documents in terms of number of words, smallest and largest class.}
\label{tab:corpora_stats}
\begin{tabular}{|l|c|c|c|c|c|}
\hline
\bf Dataset & webkb & r8 & r52 & 20ng & cade \\
\hline
\hline
\bf Classes & 4 & 8 & 52 & 20 & 12 \\
\hline
\bf Documents & 4199 & 7674 & 9100 & 18821 & 40983 \\
\hline
\bf Vocabulary & 7770 & 17387 & 19241 & 70213 & 193997 \\
\hline
\bf \vtop{\hbox{\strut Average number of}\hbox{\strut words in document}} & 909 & 390 & 418 & 851 & 913 \\
\hline
\bf Smallest class & 504 & 51 & 3 & 628 & 625 \\
\hline
\bf Largest class & 1641 & 3923 & 3923 & 999 & 8473 \\
\hline
\bf \vtop{\hbox{\strut Average size of}\hbox{\strut classes}} & 1049 & 959 & 175 & 914 & 3415 \\
\hline
\bf \vtop{\hbox{\strut Standard deviation}\hbox{\strut of sizes of classes}} & 408 & 1309 & 613 & 94 & 2451 \\
\hline
\bf \vtop{\hbox{\strut Relative}\hbox{\strut standard deviation}\hbox{\strut of sizes of classes}} & 0.39 & 1.36 & 3.51 & 0.10 & 0.72 \\
\hline
\end{tabular}
\end{table}

\subsection{Quality Measure}
\label{subsection:quality_measure}

The macro--averaged F1 score is used as a quality evaluation of the experiments' results presented in this paper.
The precision and recall for corresponding classes are calculated as follows:

\begin{equation}
    Precision(i) = \frac{tp_i}{tp_i+fp_i}\;,
    \label{eq:precision_clustering}
\end{equation}

\begin{equation}
    Recall(i) = \frac{tp_i}{tp_i+fn_i}\;,
    \label{eq:recall_clustering}
\end{equation}
where $tp_{i}$ is the number of items of class $i$ that were classified as members of class $i$, $fp_i$ is the number of items of class other than $i$ that were wrongly classified as members of class $i$ and $fn_i$ is the number of items of class $i$ wrongly classified as members of class other than $i$.
The class' F1 score is given by the following formula:

\begin{equation}
    F1(i) = 2 \cdot \frac{Precision(i)Recall(i)}{Precision(i) + Recall(i)}.
    \label{eq:fractional_F_measure}
\end{equation}

The overall quality of the classification can be obtained by taking the unweighted average F1 scores for each class. It is given by the equation:

\begin{equation}
    F1 = \frac{1}{c} \sum\limits_{i} {F1(i)},
    \label{eq:overall_F_measure}
\end{equation}
where $c$ is the number of all classes. The F1 score value ranges from 0 to 1, with a higher value indicating a higher classification quality.

The error is defined as:
\begin{equation}
    Error = 1 - F1
    \label{eq:error_measure}.
\end{equation}

The error reduction is defined as:
\begin{equation}
    ErrorReduction = \frac{(Error_{ref}-Error_{new})}{Error_{ref}},
    \label{eq:error_reduction_measure}
\end{equation}
where $Error_{ref}$ is a reference value of error and $Error_{new}$ is the new value of error.

To compare the results with other studies, micro--averaged accuracy is used. Micro--averaging does not take imbalance of classes into account.
\begin{equation}
    Accuracy = \frac{\sum tp_i}{n},
\end{equation}
where $n$ is a number of all training samples.

\subsection{Results}
\label{subsection:results}

\begin{figure*}[htb]
\centering
  \subfloat[KNN 1]{%
    \includegraphics[width=.5\hsize]{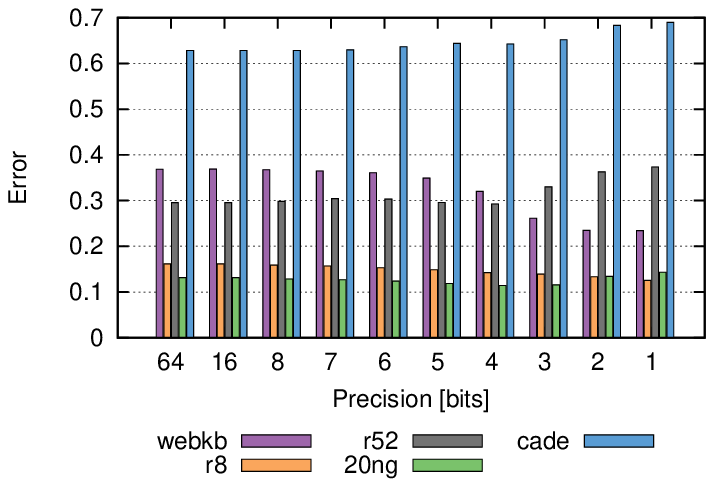}}\hfill
  \subfloat[KNN 5]{%
    \includegraphics[width=.5\hsize]{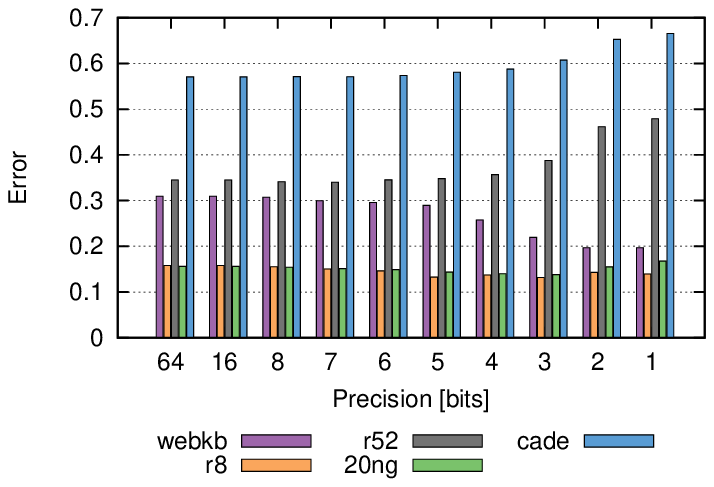}}\
  \subfloat[Logistic Regression]{%
    \includegraphics[width=.5\hsize]{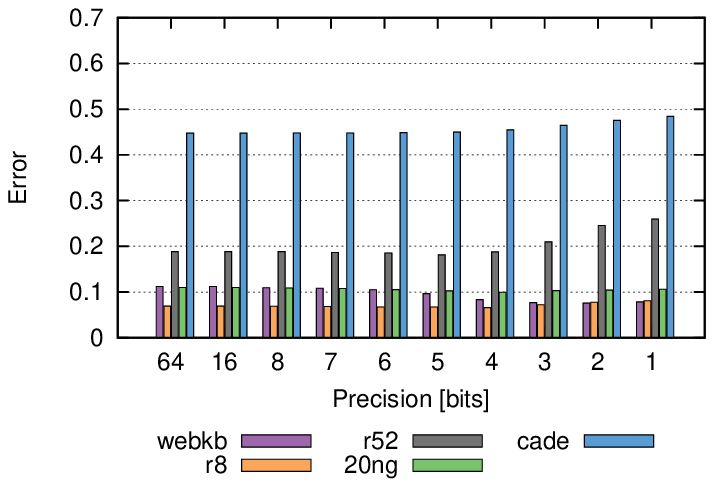}}\hfill
  \subfloat[SVM]{%
    \includegraphics[width=.5\hsize]{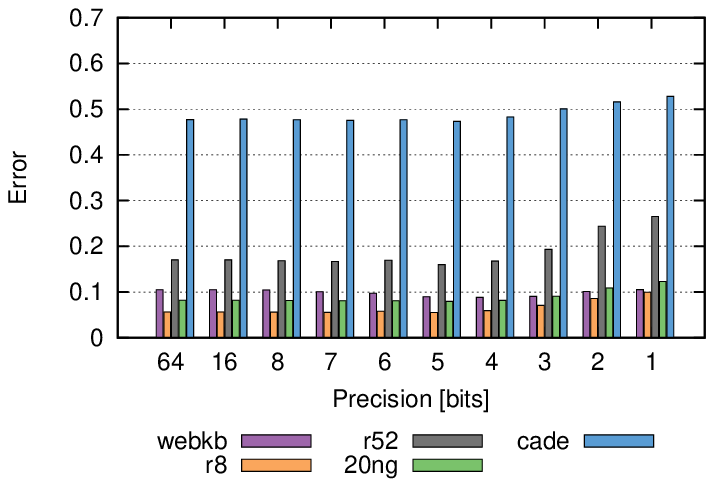}}\hfill
  \caption{Error values of the classifiers on the corpora in function of precision bits.}
\label{fig:1}
\end{figure*}

Error values on the corpora for each classifier in function of precision bits are shown in Fig. \ref{fig:1}.
For every dataset logistic regression and SVM obtain smaller error than KNNs. LR and SVM are more powerful because they model inputs (i.e. terms) in relation to classes.
Precision reduction with KNNs improves results on \texttt{webkb}, \texttt{r8} and \texttt{20ng} datasets.
KNN 5 scores higher than KNN 1 on \texttt{webkb} and \texttt{cade}.

\begin{figure}[htb]
\centering
  \subfloat[Macro--averaged]{%
    \includegraphics[width=\hsize]{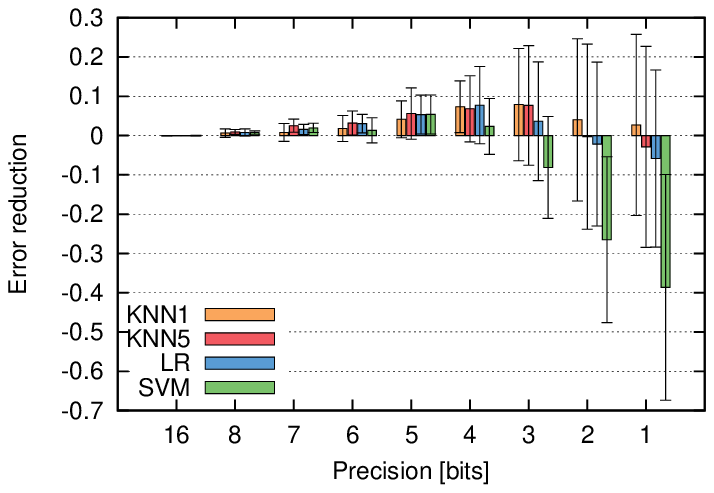}}\
  \subfloat[Micro--averaged]{%
    \includegraphics[width=\hsize]{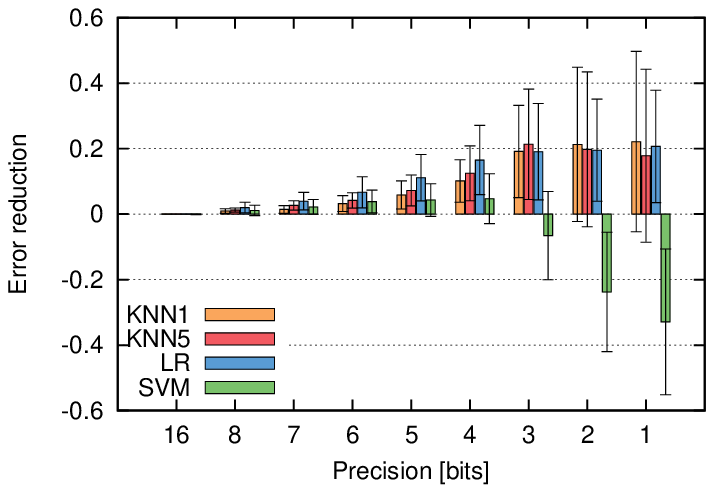}}\
  \caption{Average and standard deviation of error reduction among the corpora for the classifiers in function of precision bits.}
\label{fig:2}
\end{figure}

Fig. \ref{fig:2} shows averaged error reduction among the corpora for the classifiers.
For SVM the precision reduction is the least beneficial.
For other classifiers macro--averaged errors decrease with the reduction of precision down to 3 bits.
However, micro--averaged errors are the smallest for the precision of 1-3 bits.
Four times reduction of precision from 64 bits to 16 bits does not change the classification results.

\begin{figure}
\centering
\includegraphics[width=\hsize]{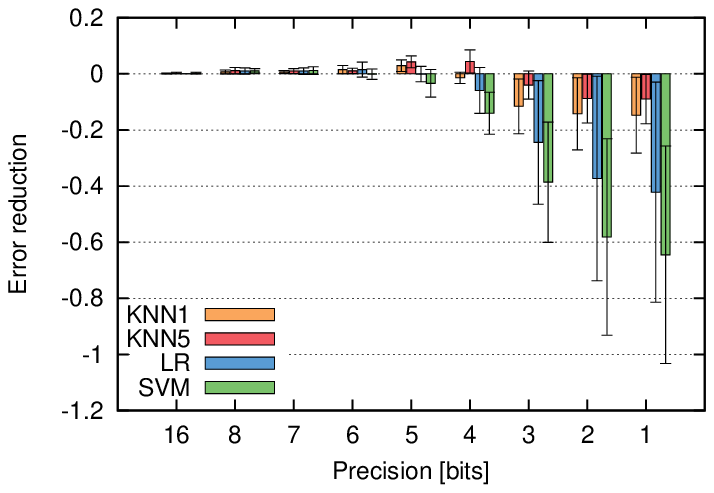}
\caption{Average and standard deviation of error reduction among the corpora for the classifiers in function of precision bits for the variant with a precision reduction after SVD.}
\label{fig:3}
\end{figure}

Fig. \ref{fig:3} shows averaged error reduction measure among the corpora for the classifiers with a precision reduction after SVD.
The results indicate that introducing the precision reduction after SVD generates more errors in every case.

\begin{figure}
\centering
\includegraphics[width=\hsize]{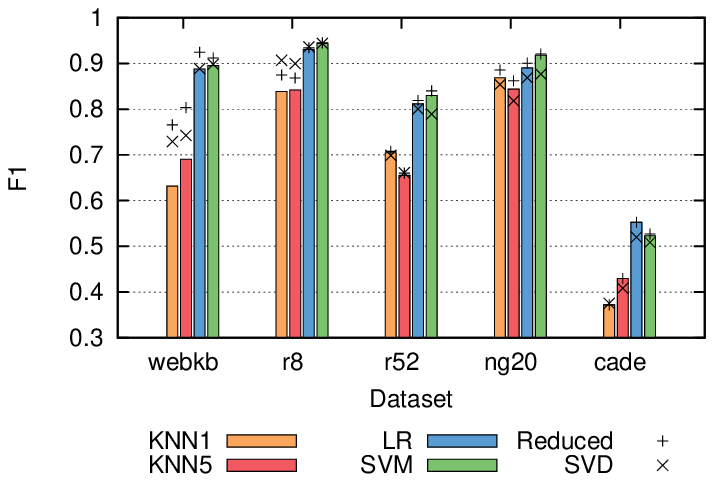}
\caption{F1 of the classifiers on the corpora with only TF--IDF, TF--IDF with the best precision reduction and TF--IDF with the best SVD.}
\label{fig:4+6}
\end{figure}

Fig. \ref{fig:4+6} presents F1 measure for 3 variants: TF--IDF, TF--IDF with the best precision reduction and TF--IDF with the best SVD.
Precision reduction gives better or similar results as applying SVD except for KNNs on \texttt{r8}.
k--nearest neighbors algorithm with precision reduction gives similar results as raw logistic regression on \texttt{r8}, \texttt{r52}, and \texttt{20ng} datasets.
In the raw form SVM has the best results for the English datasets.

\begin{figure}
\centering
\includegraphics[width=\hsize]{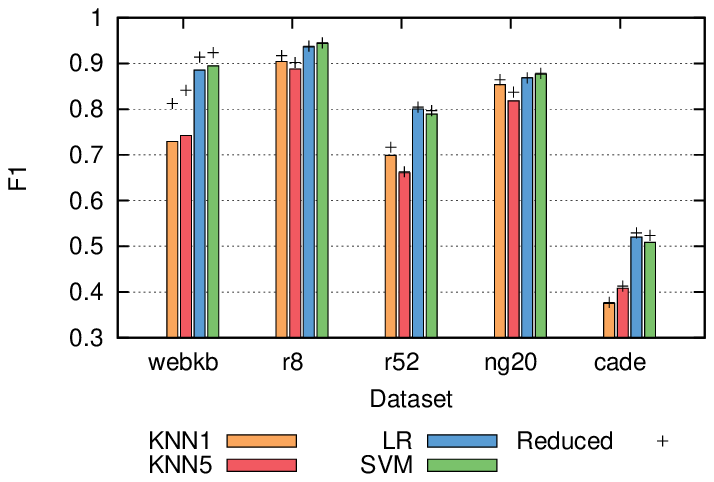}
\caption{F1 of the classifiers on the corpora with TF--IDF with the best precision reduction and SVD compared to TF--IDF with SVD.}
\label{fig:5}
\end{figure}

Fig. \ref{fig:5} presents comparison of F1 score on variant TF--IDF with SVD with and without precision reduction before SVD. Precision reduction before SVD has always positive impact, especially seen on \texttt{webkb} dataset.

Table \ref{tab:results_v} shows overall macro--averaged F1 scores for every classifier on each corpus.
The best results are obtained by logistic regression and SVM.
Classification of \texttt{cade} is the most difficult task, the best classifier has only 55\% of F1 measure.

\begin{table}
\centering
\caption{Macro--averaged F1 in 5-fold cross-validation scheme for each corpus and each classifier.}
\label{tab:results_v}
\begin{tabular}{|l|c|c|c|c|c|}
\hline
\bf  & webkb & r8 & r52 & 20ng & cade \\
\hline
\bf KNN 1 & 76.54 & 87.47 & 70.76 & 88.56 & 37.17  \\
\hline
\bf KNN 5 & 80.33 & 86.80 & 66.00 & 86.21 & 42.96  \\
\hline
\bf Logistic Regression & \textbf{92.44} & 93.41 & 81.88 & 90.04 & \textbf{55.25}  \\
\hline
\bf Linear SVM & 91.17 & \textbf{94.48} & \textbf{84.02} & \textbf{92.04} & 52.67  \\
\hline
\end{tabular}
\end{table}

Table \ref{tab:results_v2} shows overall micro--averaged accuracy for every classifier on each corpus compared with the results of SVM from \cite{2007:phd-Ana-Cardoso-Cachopo} and SVM with random search from \cite{Puurula2012}.
Our SVM with precision reduction is superior on 4 datasets: \texttt{webkb}, \texttt{r52}, \texttt{20ng} and \texttt{cade}.

\begin{table}
\centering
\caption{Micro--averaged accuracy in 5-fold cross-validation scheme for each corpus and each classifier compared to another system.}
\label{tab:results_v2}
\begin{tabular}{|l|c|c|c|c|c|}
\hline
\bf  & webkb & r8 & r52 & 20ng & cade \\
\hline
\bf KNN 1 & 80.28 & 94.81 & 90.49 & 88.67 & 41.67  \\
\hline
\bf KNN 5 & 84.30 & 94.99 & 90.28 & 86.36 & 47.47  \\
\hline
\bf Logistic Regression & 92.78 & 96.57 & 93.89 & 90.04 & 59.07  \\
\hline
\bf Linear SVM & 92.11 & 97.69 & 95.96 & 92.27 & 61.07  \\
\hline
\hline
\bf Best & \textbf{92.78} & 97.69 & \textbf{95.96} & \textbf{92.27} & \textbf{61.07}  \\
\hline
\hline
\bf SVM (\cite{2007:phd-Ana-Cardoso-Cachopo}) & 86.97 & 97.08 & 95.08 & 91.53 & 53.57 \\
\hline
\bf \vtop{\hbox{\strut SVM with}\hbox{\strut random search (\cite{Puurula2012})}} & 92.69 & \textbf{97.90} & 95.37 & 84.39 & 60.69 \\
\hline
\end{tabular}
\end{table}

Table \ref{tab:time} presents execution time of training and testing.
SVD is the most time--consuming phase in training in comparison to classification. However, it can reduce time of testing.
Time of testing using KNNs is higher than other classifiers, because it is proportional to number of documents.
Time of precision reduction is negligible.

\begin{table*}

\centering
\caption{Times of training and testing for the classifiers on the corpora without or with SVD with different number of components.}
\label{tab:time}
\begin{tabular}{|l|l||r|r|r|r|r||r|r|r|r|r|}
\cline{3-12}
 \multicolumn{2}{c|}{} & \multicolumn{5}{|c||}{\bf Training (seconds)} & \multicolumn{5}{c|}{\bf Testing (seconds)} \\
 \hline
\bf Classifier & \bf Variant & webkb & r8 & r52 & ng20 & cade & webkb & r8 & r52 & ng20 & cade \\
\hline
\hline
\multirow{4}{*}{KNN 1} & no SVD & 1.44 & 0.44 & 0.48 & 3.22 & 24.86 & 0.50 & 0.50 & 0.66 & 3.56 & 11.78 \\
 & SVD(100) & 10.48 & 10.32 & 10.23 & 17.23 & 55.77 & 0.54 & 0.38 & 0.43 & 1.54 & 9.05 \\
 & SVD(500) & 51.12 & 55.34 & 55.38 & 78.15 & 204.10 & 0.57 & 0.48 & 0.60 & 2.23 & 12.96 \\
 & SVD(1000) & 94.20 & 101.02 & 103.53 & 146.47 & 382.13 & 0.67 & 0.68 & 0.88 & 3.17 & 18.45 \\
\hline
\multirow{4}{*}{KNN 5} & no SVD & 1.44 & 0.44 & 0.48 & 3.22 & 24.86 & 0.52 & 0.57 & 0.76 & 3.99 & 13.27 \\
 & SVD(100) & 10.48 & 10.32 & 10.23 & 17.23 & 55.77 & 0.59 & 0.46 & 0.54 & 1.97 & 10.83 \\
 & SVD(500) & 51.13 & 55.34 & 55.38 & 78.15 & 204.11 & 0.65 & 0.55 & 0.70 & 2.65 & 14.80 \\
 & SVD(1000) & 94.22 & 101.02 & 103.53 & 146.47 & 382.13 & 0.74 & 0.74 & 0.98 & 3.54 & 20.32 \\
\hline
\multirow{4}{*}{Logistic Regression} & no SVD & 1.56 & 0.79 & 3.03 & 11.14 & 45.39 & 0.35 & 0.11 & 0.12 & 0.79 & 5.39 \\
 & SVD(100) & 10.57 & 10.73 & 13.18 & 22.41 & 64.44 & 0.40 & 0.14 & 0.15 & 0.83 & 5.50 \\
 & SVD(500) & 51.69 & 57.76 & 72.39 & 95.28 & 237.10 & 0.39 & 0.16 & 0.18 & 1.00 & 5.90 \\
 & SVD(1000) & 95.31 & 105.60 & 135.25 & 177.72 & 450.09 & 0.43 & 0.20 & 0.24 & 1.21 & 6.40 \\
\hline
\multirow{4}{*}{SVM} & no SVD & 1.64 & 0.72 & 1.38 & 6.78 & 38.90 & 0.35 & 0.11 & 0.12 & 0.79 & 5.39 \\
 & SVD(100) & 10.58 & 10.49 & 11.30 & 18.61 & 60.38 & 0.40 & 0.14 & 0.15 & 0.83 & 5.50 \\
 & SVD(500) & 51.66 & 56.33 & 59.63 & 84.52 & 227.25 & 0.39 & 0.16 & 0.18 & 1.00 & 5.90 \\
 & SVD(1000) & 95.37 & 103.28 & 112.41 & 159.62 & 437.49 & 0.43 & 0.20 & 0.24 & 1.21 & 6.40 \\

\hline
\end{tabular}
\end{table*}

%% file: contents-30-conclusions-future.tex
\section{Conclusions and Future Work}
\label{section:conclusions}

The conducted experiments show that it is beneficial to the perform precision
reduction on the term--document representations. However, it is unclear what number of
bits gives the best results for the specific corpus.
For some corpora, a precision reduction to 1 bit is possible without loss of accuracy.
On the other hand it is safe to reduce the number of bits from 64 to 4, which usually
improves the quality of the obtained results and never leads to their degradation.
As such, precision reduction seems the be very promising result, especially combined
with FPGA implementation, which should lead to significant computation speed-up and
memory footprint reduction.

The precision reduction is also a good alternative to dimensionality reduction by SVD.
It can lead to better accuracy and does not required time-consuming matrix manipulation. This
feature is specially important for scenarios with very large vocabularies and document data sets.
If SVD is still considered, the precision reduction should be applied before SVD, not in opposite order.

It should be also observed that focusing on micro--averaged objective allows for stronger
reduction than in macro-averaged measures.

Nowadays neural networks are one of the most popular machine learning tools used to solve NLP problems.
Our further research will be focused on testing precision reduction on distributional representations,
which are typically used as inputs to neural networks.
It is not uncommon that neural networks have millions of parameters -- the reduction of precision
of the vector weights is an interesting direction of research, which will be pursued in
our future work.